\documentclass{article}

\usepackage{amsmath,amsthm,amsfonts,amssymb}
\usepackage{algorithm}
\usepackage{algorithmic}
\usepackage[autostyle]{csquotes}
\usepackage{bm}
\usepackage{booktabs}
\usepackage{float}
\usepackage[margin=1in]{geometry}
\usepackage{tikz}
\usetikzlibrary{bayesnet}

\newcommand{\argmax}[1]{\underset{#1}{\text{arg max }}}

\newcommand{\Bern}[1]{\text{Bernoulli}(#1)}
\newcommand{\Bet}[1]{\text{Beta}(#1)}
\newcommand{\Cat}[1]{\text{Categorical}(#1)}
\newcommand{\const}{\text{ const }}
\newcommand{\Dir}[1]{\text{Dirichlet}(#1)}
\newcommand{\DKL}[2]{\text{D}_{\text{KL}}(#1 \ || \ #2)}
\newcommand{\dom}{\text{dom }}
\newcommand{\E}[1]{\underset{#1}{\mathbb{E}}}
\newcommand{\iid}{\overset{iid}{\sim}}
\newcommand{\ictr}{\mathbf{1}}
\newcommand{\limit}[1]{\underset{#1}{\text{limit }}}
\newcommand{\maximum}[1]{\underset{#1}{\text{max }}}
\newcommand{\N}{\mathbb{N}}
\newcommand{\R}{\mathbb{R}}
\newcommand{\X}{\mathcal{X}}

\usepackage{cleveref}
\crefformat{section}{\S#2#1#3}
\crefformat{subsection}{\S#2#1#3}
\crefformat{subsubsection}{\S#2#1#3}

\newtheorem*{definition}{Definition}
\newtheorem{lemma}{Lemma}
\newtheorem{theorem}{Theorem}

\usepackage[round]{natbib}

\title{Thompson Sampling for Noncompliant Bandits\thanks{Manuscript in progress. Copyright
  2019 by the author(s).}}
\author{Andrew Stirn\\ Columbia University \and 
		Tony Jebara\\ Columbia University, Netflix}

\begin{document}

\maketitle

%

%

%
%
%
%

\begin{abstract}
Thompson sampling, a Bayesian method for balancing \textit{exploration} and \textit{exploitation} in bandit problems, has theoretical guarantees and exhibits strong empirical performance in many domains. Traditional Thompson sampling, however, assumes perfect compliance, where an agent's chosen action is treated as the implemented action. This article introduces a stochastic \textit{noncompliance} model that relaxes this assumption. We prove that any \textit{noncompliance} in a 2-armed Bernoulli bandit increases existing regret bounds. With our \textit{noncompliance} model, we derive Thompson sampling variants that explicitly handle both observed and latent \textit{noncompliance}. With extensive empirical analysis, we demonstrate that our algorithms either match or outperform traditional Thompson sampling in both compliant and noncompliant environments.
\end{abstract}

\section{Introduction}

Multi-Armed Bandit (MAB) \citep{ref:sutton_barto} problems are a class of sequential decision-making problems where an agent seeks to maximize rewards by acting in an unknown stationary environment. The MAB problem is often caricaturized using a set of slot machines with unknown payout distributions. The agent must decide which arm to pull in order to maximize earnings. Because the machines' reward distributions are initially unknown, the bandit must select actions that balance exploration (learning the reward distributions) with exploitation (playing the machine with highest expected reward). Contextual bandits (CB) \citep{ref:li_CB} are a slightly modified MAB problem where the reward distributions are conditioned on an observation which is revealed to the agent prior to the selection of an action. The agent's optimal action is no longer a fixed constant but rather a function of the observed context.

Thompson sampling (TS) \citep{ref:ts} is a leading framework for implementing multi-armed bandits and contextual bandits due to its strong empirical performance across many real-world applications such as display advertising \citep{ref:chapelle_ts_empirical} and content recommendation \citep{ref:li_ts_content}. Furthermore, TS has theoretical and optimality properties (notably, regret bounds) in the non-contextual and contextual setting \citep{ref:agrawal_mab_bound_1,ref:agrawal_mab_bound_2,ref:agrawal_cb_bound,ref:korda_optimality,ref:russo_l2o,ref:russo_info_thry}. However, many TS strategies assume perfect compliance, which in many scenarios, can be fallacious and undermine learning the causal action-reward relationship. In display advertising, a bandit may select an add only to have a downstream system suppress its selection. In clinical trials, bandits can optimize health outcomes over a discrete treatment space, for example \{placebo, drug A, drug B\}. When therapies are administered by a medical professional, an attending physician may overrule the bandit in the event of a misdiagnosis or the onset of life threatening complications (i.e. an allergic reaction). When patients are responsible for administering their own treatments, noncompliance can be severe considering that patients' adherence to long-term therapies for chronic illness has been observed at 50\% in developed countries \citep{ref:nc_who}. Furthermore, noncompliance can be confounded by context (age, sex, socioeconomic status, etc.) and the willingness to comply (patients may be more averse to complying with the more aggressive treatments). Unobserved confounders and noncompliance has appeared in existing bandit literature \citep{ref:cf_bandits, ref:comp_aware, ref:iab}. Section \cref{sec:related_work} further discusses these methods and situates them relative to our proposed approach.

Thompson sampling, which is Bayesian in nature, often employs convenient Conjugate-Exponential Family (CEF) distributions that result in closed-form posteriors. Variational methods have recently been leveraged to extend TS to posteriors that cannot be computed in closed form or suffer from intractability \citep{ref:urteaga_vi_ts}. In order to explicitly model noncompliance, we will leverage tools from both CEF and variational methods

In this work, we explicate the noncompliance setting where the actions that our learned policy suggests may be overridden by another mechanism which executes a different real action that causally induces the reward. In the following sections, we demonstrate how standard TS can fail to achieve low regret in the noncompliance setting. We then derive extensions to TS to handle noncompliance. We first consider the simpler scenario where the noncompliance is observed and the ultimately executed actions are known and can be used to appropriately update the agent's policy. We then move to the more complicated scenario where the the noncompliance is latent and the executed actions must be inferred. With extensive testing, we demonstrate the performance advantages of these novel TS variants. Most notably, our variants outperform traditional TS in the presence of noncompliance and, despite their expanded models, perform similarly to TS in fully compliant environments.

\subsection{Traditional Bandit Model}
\begin{figure}[t]
\centering
\begin{tikzpicture}

	\node[obs] (x) {$x_t$};
	\node[obs, right=0.3cm of x] (a) {$a_t$};

	\node[obs, above=0.25cm of x] (r) {$r_t$};
	\node[latent, right=1.25cm of r] (mu) {$\mu$};	
	\node[const, right=4.5cm of mu] (alpha) {$\alpha$};
	
	\factor[right=2cm of mu] {mu-f} {above:Reward Prior: Varies} {alpha} {mu}; 	
			
	\edge {x} {mu};
	\edge {mu-f} {mu};
	\edge {a} {mu};
	\edge {mu} {r};
	
	\plate {p-mu} {(mu)(mu-f)(mu-f-caption)} {$x \in \X, \ K_a$};

\end{tikzpicture}
\caption{Contextual Bandit with Observed NC}
\label{fig:gm_ts}
\end{figure}
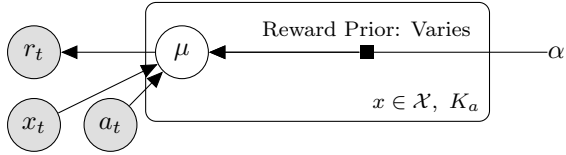

The standard CB environmental model appears in figure \ref{fig:gm_ts}. Throughout, we will use $\mu$ to denote the latent reward model parameter that has some prior $\alpha$. This model reduces to a MAB environment when $|\mathcal{X}|=1$. There exists environmental causality wherever a directed edge exists. We say \enquote{environmental} causality to distinguish from the agent's policy that maps context to action. Shaded nodes represent observable variables, while transparent nodes are latent. Note there are $K_a \in \N_{++}$ tiles such that conditioning $\mu$ on $a_t$ is an indexing operation. The same is true for discrete context ($|\X|<\infty$). When context is continuous, conditioning $\mu$ on $x_t$ requires a conditional probability. Choosing CEF distributions to model the environment can result in convenient conditional and posterior probabilities.
Regret for this model is typically assessed w.r.t. the true reward distribution $\mu$ and the agent's chosen control setting $a_t$ (equation \ref{eq:regret_fc}). We will have to reconsider this notion of regret when we introduce noncompliance in \cref{sec:nc}.
\begin{equation}
\label{eq:regret_fc}
\begin{aligned}
	\mathcal{R} &= \sum_{t=1}^T \Big[
		\maximum{a^*} \E{}[r|x_t,a^*,\mu] - \E{}[r|x_t,a_t,\mu]
		\Big]
\end{aligned}
\end{equation}

\subsection{Thompson Sampling}
\label{sec:ts}
It is informative to recall that TS is fundamentally the application of Bayesian inference in a sequential manner. It maintains a posterior distribution over unknowns such as reward model parameters (and potentially any other latent variables) given all observations that have been collected so far. Initially, with no data, the posterior is merely the prior distribution which is often initialized to have high entropy to encourage exploration. One advantage of TS is the ability to employ priors that encode a priori assumptions about the environment. In fact, some work has shown that the use of non-informative priors for multi-parameter Gaussian bandits can be risky \citep{ref:priors}. This article only considers non-informative priors for Bernoulli bandits. Even as some data is collected, the posterior continues to reflect the uncertainty in model parameters due to the limited exploration that has been available so far. To collect more data, TS samples the posterior, obtains a proposed model, and then acts greedily w.r.t. the model to maximize reward. As TS collects more data, the observations begin to overwhelm the prior and the posterior converges to a more peaked distribution over model parameters. At that point, the models it samples are less diverse and it has essentially committed to a small subset of models which it uses to act greedily. The posterior's transition from a high entropy prior to a low entropy distribution over parameters is precisely how TS manages its progression from exploration to exploitation. Remarkably, this sequential application of Bayesian inference has strong theoretical and empirical performance. The ingredients for this Bayesian recipe are:
\begin{enumerate}
	\setlength\itemsep{-2pt}
	\item A set of a priori observable variables $X_t$
	\item A set of controllable variables $A_t$
	\item A set of a posteriori observable variables $R_t$ that includes the reward signal $r_t$
	\item A probabilistic model that defines the set of latent parameter(s) $\theta$ we wish to learn
	\item A data buffer: $\mathcal{D}=\{(X_t, A_t, R_t)\}_{t=1}^{T}$
	\item An i.i.d. reward signal likelihood $p(r_t \ | \ X_t, A_t, \theta)$
	\item Prior(s) for the latent model parameter(s) $p(\theta)$
	\item A sampleable posterior (or its approximation) $q(\theta) \approx p(\theta \ | \ \mathcal{D}) \propto p(\mathcal{D} \ | \ \theta)p(\theta).$
\end{enumerate}
We depart from the traditional (context, action, reward) tuples for the sake of generality. While perhaps notationally cumbersome, our modification to (a priori observations, control settings, a posteriori observations) tuples is a generalization that reduces to MAB and CB and allows us to consider expanded models that capture additional mechanisms such as noncompliance. A MAB scenario occurs when $X_t$ is the empty set, $A_t$ contains the agent's chosen action, and $R_t$ contains the reward signal. A CB occurs identically except that $X_t$ is no longer empty and instead contains the context revealed to the agent prior to action selection.

TS balances exploration and exploitation by setting controllable variables to $A'$ with the probability expressed in equation \ref{eq:ts_prob}. However, one does not need to compute these probabilities but can rather sample $\theta$ from its posterior, $p(\theta \ | \ \mathcal{D})$, and act greedily with respect to $\E{}[r \ | \ X, A, \theta]$. Algorithm \ref{alg:ts} details this sampling mechanism.
{\scriptsize
\begin{equation}
\label{eq:ts_prob}
\begin{aligned}
	 \int \ictr \Bigg[
		\E{p(r | \theta, \mathcal{D})}\Big[r | X,A',\theta\Big] = 
		\maximum{A} \E{p(r | \theta, \mathcal{D})}\Big[r | X,A,\theta\Big]
	\Bigg] p(\theta | \mathcal{D}) d\theta
\end{aligned}
\end{equation}}
\begin{algorithm}[H]
   \caption{Thompson Sampling Framework}
   \label{alg:ts}
\begin{algorithmic}
	\REQUIRE Parameters for $p(\theta)$ 
	\STATE Initialize data buffer $D = \{\emptyset\}$
	\FOR{$t \in \{1, \hdots, T\}$}
		\STATE Environment reveals a priori context variables $X_t$
		\STATE Agent samples $\theta \sim p(\theta \ | \ X_t, \mathcal{D})$ (or approx.  $q(\theta)$)
		\STATE Agent sets control to maximize expected reward\\
		$\ \ \ A_t = \argmax{A} \E{p(r|X_t,A,\theta)}[r|X_t,A,\theta]$
		\STATE Env. reveals a posteriori variables and reward $R_t$
		\STATE Data buffer update $\mathcal{D} := \mathcal{D} \cup {(X_t, A_t, R_t)}$
		\STATE Posterior update $p(\theta \ | \ \mathcal{D})$ (or its approx. $q(\theta)$)
	\ENDFOR
\end{algorithmic}
\end{algorithm}

\section{Noncompliance}
\label{sec:nc}
Perhaps there are various notions of noncompliance in bandit systems. This article considers the following rigid definition for a noncompliant environment.
\begin{definition}
If at anytime an agent's chosen action within an environment is not unequivocally the implemented action, then the environment is noncompliant.
\end{definition}

\subsection{Modeling Noncompliance}
\label{sec:nc:mdl}
\begin{figure}[t]
\centering
\begin{tikzpicture}

	\node[obs] (x) {$x_t$};
	\node[obs, right=0.3cm of x] (a) {$a_t$};

	\node[obs, above=0.25cm of x] (r) {$r_t$};
	\node[latent, right=1.25cm of r] (mu) {$\mu$};
	\node[const, right=4.5cm of mu] (alpha) {$\alpha$};

	\node[obs, below=0.25cm of x] (z) {$z_t$};
	\node[latent, right=1.25cm of z] (pi) {$\pi$};
	\node[const, right=4.5cm of pi] (beta) {$\beta$};

	\factor[right=2cm of mu] {mu-f} {above:Reward Prior: Varies} {alpha} {mu};
	\factor[right=2cm of pi] {pi-f} {above:Compliance Prior: Dirichlet} {beta} {pi};

	\edge {x} {mu, pi};
	\edge {mu-f} {mu};
	\edge {a} {mu};
	\edge {mu} {r};
	\edge {pi} {a};
	\edge {z} {pi};

	\plate {p-mu} {(mu)(mu-f)(mu-f-caption)} {$x \in \X, \ K_a$};
	\plate {p-pi} {(pi)(pi-f)(pi-f-caption)} {$x \in \X, \ K_z$};

\end{tikzpicture}
\caption{Contextual Bandit with Observed NC}
\label{fig:gm_obs}
\end{figure}
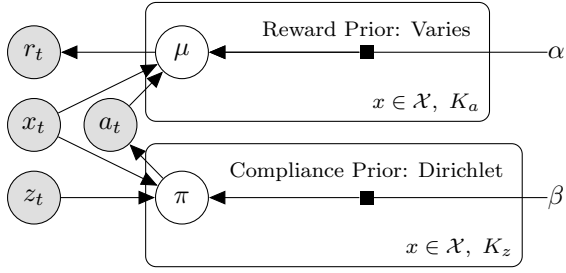

We seek a model that is adaptable to both compliant and noncompliant environments. Following the TS Bayesian recipe, we place the agent's proposed action $z_t \in A_t$ (control variables). Departing from the MAB and CB, we place a tuple of the environment's implemented action and reward signal $(a_t, r_t) \in R_t$ (the a posteriori variable set). Naturally, we place $x_t \in X_t$ for CB, whereas $X_t = \{\emptyset\}$ for MAB. Because proposed actions $z_t$ and implemented actions $a_t$ are discrete, we use a Categorical distribution to model noncompliance with parameters $\pi$, upon which we place a conjugate Dirichlet prior with parameters $\beta$.
\begin{equation*}
\label{eq:nc_mdl}
\begin{aligned}
	\pi|x,z &\iid \Dir{\beta}\\
	\pi|x,z &\in \{\Re^{K_a} : ||\pi||_1 = 1, \ \pi \succeq 0 \}\\
	a_t &\iid \Cat{\pi | x_t, z_t} \\
\end{aligned}
\end{equation*}
In figure \ref{fig:gm_obs}, we show the corresponding graphical model for a CB with observed noncompliance (implemented actions are observable). This model supports a different number of proposed actions $K_z$ than implemented actions $K_a$. However, we generally consider the case when $K_z = K_a = K$. We additionally assume that context may have a causal influence on noncompliance. Once again, imagine the probability a patient complies with a prescribed therapy as potentially being a function of context. Continuing with the example, the observability of $a_t$ corresponds to all patients honestly reporting their implemented actions.
This model reduces to the standard CB when $K_z = K_a$ and $(\pi | x_t, z_t) = e_{z_t}$, where $e_{z_t}$ is a basis vector containing all zeros except a single 1 in the $z_t^{th}$ position.

In our noncompliance model, implemented actions are no longer under the agent's direct control. Thus, we feel obliged to recompile our notion of regret to be w.r.t. that which the agent has control over. To do so, we condition the expected reward on context $x$, proposed action $z$, and the model parameters.
\begin{equation}
\label{eq:E_r}
\begin{aligned}
	\E{}[r|x,z,\mu,\pi]
		&= \int r \sum_{a} p(r, a  | x, z, \mu, \pi) dr\\
		&= \int r \sum_{a} p(r  | x, a, \mu) p(a | x, z, \pi) dr
\end{aligned}
\end{equation}
Evaluating equation \ref{eq:E_r} with the environments true parameters, $\mu$ and $\pi$, we compute cumulative regret as:
\begin{equation}
\label{eq:regret_nc}
\begin{aligned}
	\mathcal{R} &= \sum_{t=1}^T \Big[
		\maximum{z^*} \E{}[r|x_t,z^*,\mu,\pi] - \E{}[r|x_t,z_t,\mu,\pi]
		\Big]
\end{aligned}
\end{equation}

\subsection{Related Work}
\label{sec:related_work}
\cite{ref:cf_bandits} consider unobserved confounding environmental context variables that both determine the optimal arm and a player's proclivity for selecting certain actions. While different from our notion of compliance, they do examine interference on the action space. They depart from traditional notions of regret by defining the optimal action as the one that maximizes the expected reward \textbf{conditioned on} the unobservable context using counter-factual reasoning. Another article \citep{ref:comp_aware} claims that \cite{ref:cf_bandits}'s use of the $do(\cdot)$ operator is not always realizable in real-world bandit applications.

\cite{ref:comp_aware} consider a similar notion of observed noncompliance as presented herein. They introduce three reward protocols: \textit{chosen}, \textit{actual}, and \textit{comply}. \textit{Chosen} assigns rewards to the agent's proposed action, which they claim obeys existing bounds. \textit{Actual} assigns rewards to the environment's implemented action, which they claim can outperform \textit{actual} and \textit{comply} but also fail entirely. \textit{Comply} assigns rewards only when proposed and implemented actions are the same, which they too claim can outperform the other two protocols but is unbounded. They construct and validate a \textit{hierarchical bandit} where a master bandit optimizes over three subsidiary bandits that each employ a different one of their reward protocols. Additionally, they construct and validate a TS variant that leverages this \textit{hierarchical bandit}.

\cite{ref:iab} argues that different notions of regret must be considered in the presence of noncompliance. Their definition of \textit{Intent to Treat} (ITT) is analagous to \cite{ref:comp_aware}'s \textit{chosen} reward protocol. They argue ITT regret ignores the causal effect of treatments whereas \textit{Compliers} regret (over the examples where proposed and implemented actions agree) more effectively captures this causality. They construct and validate UCB and two-stage-least-squares based bandit algorithms.

Comparing these works to ours, there are clear similarities and differences. Our regret (equation \ref{eq:regret_nc}) is identical to \cite{ref:iab}'s ITT regret. \cite{ref:comp_aware} and \cite{ref:iab} consider modified notions of regret from which they construct new algorithms. Instead, we take a model-first approach. In \cref{sec:nc:mdl}, we defined a probabilistic graphical model that captures the confounder (noncompliance) we seek to disambiguate. From here, we simply follow the Bayesian recipe (\cref{sec:ts}), which leads to the natural TS solutions we present in \cref{sec:alg}. Unlike these other works, our model-first approach allows us to consider unobserved noncompliance, which to the best of our knowledge has not been done, by treating the latent implemented actions $a_t$ variationally and using posterior approximations.

\subsection{Noncompliance Regret Bounds}
\label{sec:theory}
In this section, we consider the effect ignoring noncompliance has on existing TS regret bounds. \cite{ref:agrawal_mab_bound_2} formulated problem-dependent TS regret bounds for Bernoulli bandits (equation \ref{eq:ts_bound}). Without loss of generality, they assume that the reward parameters are ordered $\mu \in \{(0,1)^K : \mu_1 \geq \mu_i\}$.
\begin{equation}
\label{eq:ts_bound}
\begin{aligned}
    &\E{} [\mathcal{R}(T,\mu)] \leq
        (1+\epsilon) \sum_{i=2}^K \Big[f(\mu_1, \mu_i) \ln T  \Big] +
        \mathcal{O}\Big(\frac{K}{\epsilon}\Big)
\end{aligned}
\end{equation}
In this article we consider the impact noncompliance has on $f(\mu_1, \mu_i)$.
\begin{equation}
\label{eq:ts_bound_f}
\begin{aligned}
	f(\mu_1, \mu_i)
		&= \frac{ (\mu_1 - \mu_i)}
			{\mu_i\log\frac{\mu_i}{\mu_1} + (1-\mu_i)\log\frac{(1-\mu_i)}{(1-\mu_1)}}\\
	\dom f &= \{\mu \in (0,1)^2 : \mu_1 \geq \mu_i\}\\
\end{aligned}
\end{equation}
Note, we define: $f(x,x) = \frac{0}{0} = 0$
\begin{lemma}
\label{lemma:grad}
The function $f : \mu \rightarrow \R_+$ (equation \ref{eq:ts_bound_f}) is monotonic over its domain in that:
{\normalfont
\begin{equation*}
\begin{aligned}
	\frac{\partial}{\partial \mu_1}f(\mu) &\leq 0 \ \forall \ \mu \in \text{dom } f\\
	\frac{\partial}{\partial \mu_i}f(\mu) &\geq 0 \ \forall \ \mu \in \text{dom } f
\end{aligned}
\end{equation*}}
\end{lemma}
We prove lemma \ref{lemma:grad} in \cref{sec:appendix:lemma1_proof} of our appendix. Intuitively, this lemma captures the upward trend of $f$ as $\mu_1$ and $\mu_i$ move towards one another, which is precisely the concern when ignoring noncompliance.

Consider a $K$-armed Bernoulli bandit in a noncompliant environment. With absent context, collapsing figure \ref{fig:gm_obs}'s tiles allows us to represent the MAB model parameters as vectors and matrices:
\begin{equation*}
\begin{aligned}
	\mu &\in \{\R^K : 0 \preceq \mu \preceq 1\}\ \\
	\Pi &\in \{\Re^{K \times K} : \Pi \cdot \bm{1} = \bm{1}, \ \Pi_{ij} \geq 0 \}\\
\end{aligned}
\end{equation*}
We then can express the vector of expected rewards (equation \ref{eq:E_r}) for a Bernoulli MAB as:
\begin{equation*}
\begin{aligned}
	\mu' = \begin{bmatrix}
		E{}[r|z=1,\mu,\pi]\\
		\vdots\\
		E{}[r|z=K,\mu,\pi]
	\end{bmatrix}
	= \Pi \cdot \mu
\end{aligned}
\end{equation*}
\begin{theorem}
\label{thm:nc_2}
For a 2-armed bandit with any reward parameter $\mu \in \{\R^2: 0 \prec \mu \prec 1\}$ and any compliance parameter $\Pi \in \{\Re^{2 \times 2} : \Pi \cdot \bm{1} = \bm{1}, \ \Pi_{ij} \geq 0 \}$, Thompson sampling's expected regret bound (equation \ref{eq:ts_bound}) obeys $\E{} [\mathcal{R}(T,\mu)] \leq \E{} [\mathcal{R}(T,\Pi \cdot \mu)]$.
\end{theorem}
\paragraph{Proof}
Let $\mu' = \Pi \cdot \mu$. Without loss of generality, assume $\mu_1 > \mu_2$ and $\mu'_1 > \mu'_2$. The latter simply allows a permutation after computing $\mu'$. Because $\mu'_1$ and $\mu'_2$ are both convex combinations of $\mu_1$ and $\mu_2$, $\max \mu' \leq \max \mu$ and $\min \mu' \geq \min \mu$. Because noncompliance shifts the reward model (as observed by traditional TS that ignores noncompliance) from $\mu\rightarrow\mu'$, we can apply lemma \ref{lemma:grad} to say $f(\mu_1, \mu_2) \leq f(\mu'_1, \mu'_2)$. Finally and as a result of our $K=2$ assumption, there are no other $f$ terms appearing in equation \ref{eq:ts_bound}. Thus, it must be $\E{} [\mathcal{R}(T,\mu)] \leq \E{} [\mathcal{R}(T,\mu')] = \E{} [\mathcal{R}(T,\Pi \cdot \mu)]$.

One might argue that 2-armed bandits are not practically useful. To the contrary, consider a medical trial with action space \{placebo, new drug\}. Patients who were prescribed the new therapy and failed to comply can greatly confound the drug's causal effect. In this scenario, traditional TS could incur significantly more regret than an algorithm that specifically considers patient compliance.

Define the change in TS regret bounds due to ignored noncompliance as:
\begin{equation*}
\label{eq:delta_regret}
\begin{aligned}
	\Delta(T,\mu,\Pi)
		& = \E{} [\mathcal{R}(T,\Pi \cdot \mu)] - \E{} [\mathcal{R}(T,\mu)]
\end{aligned}
\end{equation*}
\begin{theorem}
\label{thm:nc_k}
For a $(K>2)$-armed bandit with reward parameter $\mu \in \{\R^K: 0 \prec \mu \prec 1\}$, simultaneously
{\normalfont
\begin{equation*}
\begin{aligned}
	&\exists \ \Pi \in \{\Re^{K \times K} : \Pi \cdot \bm{1} = \bm{1}, \ \Pi_{ij} \geq 0 \}
		: \Delta(T,\mu,\Pi) > 0\\
	&\exists \ \Pi \in \{\Re^{K \times K} : \Pi \cdot \bm{1} = \bm{1}, \ \Pi_{ij} \geq 0 \}
		: \Delta(T,\mu,\Pi) < 0\\
\end{aligned}
\end{equation*}}
\end{theorem}

\paragraph{Proof} For any $\mu \in \{\R^K: 0 \prec \mu \prec 1\}$, there exists a low-rank permutation matrix that results in $\mu' = [\max \mu, \min \mu, \hdots, \min \mu]^T$ such that $f(\mu_1, \mu_i) \leq f(\mu'_1, \mu'_i) \ \forall \ i \in [2,K]\cap\N$. For that same $\mu$ there simultaneously exists a low-rank permutation matrix that results in $\mu' = [\max \mu, \max \mu, \hdots, \max \mu]^T$ such that $f(\mu_1, \mu_i) \geq f(\mu'_1, \mu'_i) = 0 \ \forall \ i \in [2,K]\cap\N$.

\section{Algorithms for Noncompliance}
\label{sec:alg}
This section presents four TS algorithms for Bernoulli MAB. These algorithms all follow the general TS framework defined in algorithm \ref{alg:ts}. What makes these algorithms unique from one another are their underlying modeling assumptions and posterior update procedures. Concretely, the environment is specified as:
\begin{itemize}
	\setlength\itemsep{-2pt}
	\item Reward model $r_t \iid \Bern{\mu | a_t}$
	\item Noncompliance model $a_t \iid \Cat{\pi|z_t}$
	\item Reward prior $\mu | a \iid \Bet{\alpha}$
	\item Noncompliance prior $\pi | z \iid \Dir{\beta}$
\end{itemize}
Clearly, the joint likelihood factorizes according to
\begin{equation}
\label{eq:mab_nc_mdl}
\begin{aligned}
	p(r, \mu, a, \pi | z)
		&= p(r | \mu, a) p(\mu | a)p(a | \pi, z)p(\pi | z)\\
\end{aligned}
\end{equation}

\subsection{Thompson Sampling}
Standard TS ignores noncompliance by treating the proposed action $z_t$ as the implemented action $a_t$. The corresponding likelihood, $p(r, \mu | z) = p(r | \mu, z) p(\mu | z)$, does not accurately reflect the environment's true model. This model places $z_t \in A_t$ and $r_t \in R_t$. Because the reward model and its prior are conjugate, its posterior update in algorithm \ref{alg:ts} would be:
\begin{equation*}
\label{eq:ts_r_post}
\begin{aligned}
	p(\mu|z,\mathcal{D})
		&\propto p(\mathcal{D} | \mu) p(\mu|z)
		= \Bet{\alpha_{0,z}', \alpha_{1,z}'}\\
	\alpha_{r,z}' &= \alpha + \sum_{t=1}^{T} \ictr[r_t = r]\ictr[z_t = z]
\end{aligned}
\end{equation*}

\subsection{Compliance Checking}
\label{sec:ts_check}
TS with compliance checking (TS-Check) simply implements \cite{ref:comp_aware}'s \textit{comply} reward protocol. This model places $z_t \in A_t$ and $(a_t, r_t) \in R_t$, but employs the same under-specified model as TS. Rather than using compliance information probabilistically, this model uses compliance heuristically in that the posterior update only considers samples where proposed and implemented actions agree. Concretely, its posterior update in algorithm \ref{alg:ts} would be:
\begin{equation*}
\label{eq:ts_check_post}
\begin{aligned}
	p(\mu|z,\mathcal{D}) &= \Bet{\alpha_{0,z}', \alpha_{1,z}'}\\
	\alpha_{r,z}' &= \alpha + \sum_{t=1}^{T} \ictr[r_t = r]\ictr[z_t = a_t = z]
\end{aligned}
\end{equation*}

\subsection{Observed Compliance}
\label{sec:ts_obs}
TS with observed compliance (TS-Obs) places $z_t \in A_t$ and $(a_t, r_t) \in R_t$ and employs the expanded probabilistic model (equation \ref{eq:mab_nc_mdl}). The posterior factorizes $p(\mu,\pi | a, z, \mathcal{D}) =  p(\mu | a, \mathcal{D})p(\pi | z, \mathcal{D})$ such that its update in algorithm \ref{alg:ts} would be:
\begin{equation*}
\label{eq:ts_c_post}
\begin{aligned}
	p(\mu|a,\mathcal{D}) &= \Bet{\alpha_{0,z}', \alpha_{1,z}'}\\
	\alpha_{r,a}' &= \alpha + \sum_{t=1}^{T} \ictr[r_t = r]\ictr[a_t = a]\\
	p(\pi | z, \mathcal{D}) &= \Dir{\beta_{1,z}', \hdots, \beta_{K,z}'}\\
	\beta_{k,z}' &= \beta + \sum_{t=1}^{T} \ictr[a_t = k]\ictr[z_t = z]\\
\end{aligned}
\end{equation*}

\subsection{Latent Compliance}
\label{sec:ts_lat}
TS with latent compliance (TS-Lat) is a more challenging scenario where the algorithm must infer which actions were implemented across all time steps. This model corresponds to figure \ref{fig:gm_obs} except that the $a_t$ node would be transparent (see figure \ref{fig:gm_lat} of the appendix). This model places $z_t \in A_t$ and $r_t \in R_t$. With latent $a_1, \hdots, a_T$, we again follow the Bayesian procedure of maintaining distributions over all latent variables. The posterior for this model, $p(a, \mu, \pi| z, \mathcal{D})$, has no closed-form solution due to the impact latent $a_1, \hdots, a_T$ have on the probability model. Without an available closed-form posterior, we must employ variational Bayesian methods. We chose Variational Inference \citep{ref:vi, ref:vi_exp}, which employs the following mean-field posterior approximation
\begin{equation*}
\label{eq:vi_factor}
\begin{aligned}
	p(a, \mu, \pi| z, \mathcal{D})
		&\approx q(a, \mu, \pi)\\
		&=
		\Bigg[\prod_{i=1}^N q(a_i)\Bigg]
		\Bigg[\prod_{j=1}^K q(\mu_j)\Bigg]
		\Bigg[\prod_{k=1}^K q(\pi_k)\Bigg]
\end{aligned}
\end{equation*}
where the distributions being multiplied are $q(a_i) = \Cat{\phi_i(1), \hdots, \phi_i(K)}$, $q(\mu_j) = \Bet{\alpha'_{1,j}, \alpha'_{0,j}}$ and $q(\pi_k) = \Dir{\beta'_{k,1}, \hdots, \beta'_{k,K}}$. Please refer to \cref{sec:vi_deriv} of the appendix for a full derivation of these $q$ distributions and their updates. In this latent noncompliance setting, the posterior update in algorithm \ref{alg:ts} becomes a full run of the Variational Inference algorithm that we present in algorithm \ref{alg:vi} of the appendix. Note that the TS framework, as applied to our noncompliance model, only requires samples of $\mu$ and $\pi$ in order greedily propose an action. As such, TS-Lat's posterior sampling becomes:
\begin{equation*}
\begin{aligned}
	(\mu|a_t = j) &\sim q(\mu_j)\\
	(\pi|z_t = j) &\sim q(\pi_j)\\
\end{aligned}
\end{equation*}
It is important to consider the implications of this variational approach. While the underlying model is correct, which is important as demonstrated by our simulations in \cref{sec:sim}, there exists an identifiability problem. Consider the reward parameter vector $\mu' \in \{\R^K : 0 \preceq \mu' \preceq 1\}$ as that which is observable by traditional TS. We showed that our noncompliance model factorizes $\mu' = \Pi\cdot\mu$. In the observed setting, these factors' updates are intuitively disjoint--$(a_t,z_t)$ tuples determine updates for $\Pi$ and $(r_t,a_t)$ tuples determine updates for $\mu$--such that our estimates will converge to the underlying environment's true parameters. In the latent setting, however, these updates are no longer disjoint. Essentially, this latent treatment overparameterizes the observable reward parameter vector into a matrix-vector product. Consequently, there are no guarantees that we will converge to the environment's true parameters. In other words, there exists multiple local optima. Overparameterization in linear and neural networks is an active research area that examines overparameterization's effect on the optimization process, specifically the speed of convergence  \citep{ref:op_1, ref:op_2, ref:op_3}.

When developing TS-Lat, we noticed it was prone to long tail failures (high regret). To ameliorate this failure mode, we introduced a soft-start mechanism that works for MAB and CB with discrete contex. For each context realization, we prohibit variational posterior updates until $M$ samples have been collected. This modification amounts to uniform exploration for the first $M$ samples of a context realization. Setting $M = 0$ deactivates this mechanism. When referring to a specific instance of this algorithm, we will use TS-Lat-$M$ to denote the instance's soft-start parameter.

\section{Simulations}
\label{sec:sim}
In this section, we subject the algorithms described in \cref{sec:alg} to a battery of simulations in various noncompliant Bernoulli reward environments. In all simulations, we employ non-informative and, as is the case for our Bernoulli noncompliant bandit model, uniform (over the support) priors. Concretely, $\alpha = \beta = 1$.

\subsection{Regret Bound Analysis}
\label{sec:sim:mab}
\begin{figure}[ht]
\centering
\includegraphics[scale=0.4]{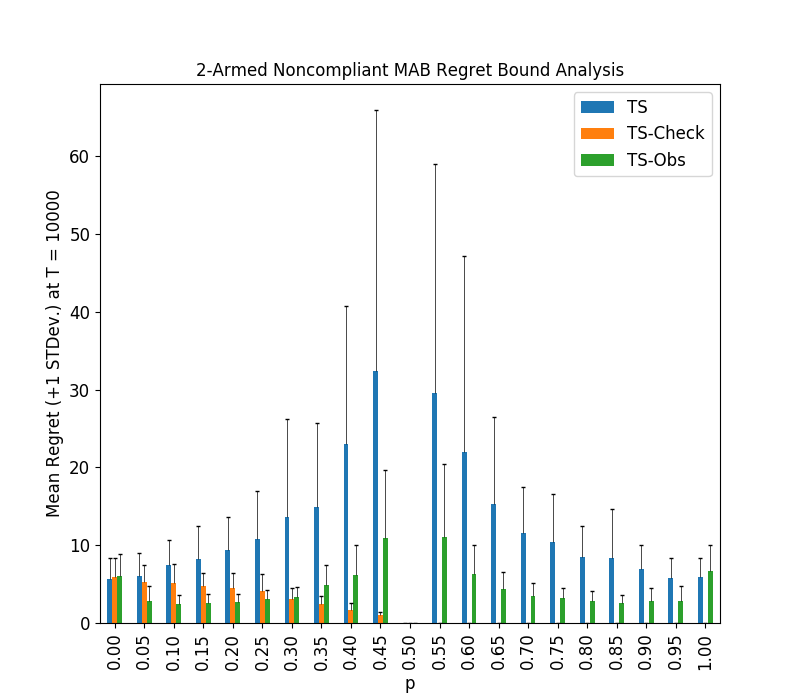}
\caption{Empirical Regret Bound Analysis}
\label{fig:bounds}
\end{figure}

This simulation seeks to empirically confirm theorem \ref{thm:nc_2} and examine how TS-Check and TS-Obs perform relative to TS in a 2-armed Benroulli bandit environment with varying levels of observable noncompliance. Specifically, we consider the following nomcompliant MAB environment:
\begin{equation*}
\label{eq:mab_env}
\begin{aligned}
    \underbrace{\begin{bmatrix} \mu'_1\\ \mu'_2\\ \end{bmatrix}}_{\mu'} &=
        \underbrace{\begin{bmatrix} (1-p) & p \\ p & (1-p)\\ \end{bmatrix}}_{\Pi}
        \underbrace{\begin{bmatrix} 0.75\\ 0.25\\ \end{bmatrix}}_{\mu}\\
\end{aligned}
\end{equation*}
In this simulation, we swept $p$ from $0\rightarrow 1$ in increments of $0.05$. For each value of $p$, we ran TS, TS-Obs, and TS-Check $100$ times each with a $T=1\text{e}4$ time horizon. We present the mean regret in figure \ref{fig:bounds} with one standard deviation error bars. Regret at $p=0.5$ is zero, as all arms are optimal. We do not plot results for TS-Check for $p>0.5$. In this regime, TS-Check incurs substantially larger regret than the other algorithms as its updates are severely suppressed. For some intuition, consider that at $p=1$ TS-Check's posterior will always be the prior. At this extreme, TS-Check chooses arms uniformally at random and therefore incurs linear regret.

Per theorem \ref{thm:nc_2}, we expect and see that TS obeys $\limit{p\rightarrow0.5} \ \E{} [\mathcal{R}(T,\mu')] = \infty$. Additionally, we see an increase in regret variance as noncompliance obfuscates, from the perspective of TS, the causal reward parameters. TS-Check clearly matches or outperforms TS in the $p \in [0,0.5)$ regime and interestingly trends downward as $p\rightarrow0.5$. We therefore propose TS-Check as a suitable solution for noncompliant MAB environments so long as the expected outcome induced by $\Pi$ is compliance (i.e. $\E{}[a|z,\pi] = z$). TS-Obs matches or outperforms TS in both mean and variance across the entire noncompliance spectrum, making it a suitable solution regardless of any compliance assumptions. In the case of full compliance ($p=0$), TS-Obs does not seem to incur any significant regret penalty despite its expanded model. From an empirical perspective, TS-Obs's factorized model (separate compliance and reward parameters) obeys an expected regret mechanism seemingly different from \cite{ref:agrawal_mab_bound_2}'s problem dependent expected regret bounds. Interestingly, TS-Obs's expected regret bound (if one exists) does not empirically appear monotonic in $p\in [0,0.5)$ and $p\in (0.5,1]$. We consider it an open problem to formulate similarly rigorous bounds for our TS-Obs algorithm.

\subsection{Contextual Bandit Simulations}
This section's simulations look to empirically measure the performance of our latent noncompliance solution, TS-Lat, against traditional TS, which assumes perfect compliance, as well as TS-Check and TS-Obs that utilize observed compliance outcomes. The secondary focus is to introduce discrete context ($|\X|=2$) where the probability of observing each context is $p(x=0)=p(x=1)=0.5$. In the case of discrete context, the CB is simply comprised of $|\X|$ MAB instances. We consider the four different noncompliance model parameters outlined in table \ref{tab:env_nc}. For each environment, we fixed the reward model parameters to:
\begin{equation*}
\begin{aligned}
	(\mu \ | \ x=0) &= \begin{bmatrix} .65 &.35 \end{bmatrix}^T\\
	(\mu \ | \ x=1) &= \begin{bmatrix} .25 &.75 \end{bmatrix}^T
\end{aligned}
\end{equation*}
We tried soft-start settings of $M\in\{0,40\}$. For Bernoulli rewards with noninformative priors, the soft-start effectively samples the prior until it begins updating a context's posterior after collecting $M$ samples.
\begin{table}[ht]
	\centering
	\small
  	\begin{tabular}{ccccc}
    		\toprule
		Env. &1 &2 &3 &4\\
		\midrule
		$\Pi \ | \ x=0$
			&$\begin{bmatrix} 1 &0 \\ 0 &1\end{bmatrix}$
			&$\begin{bmatrix} 0 &1 \\ 1 &0\end{bmatrix}$
			&$\begin{bmatrix} .7 &.3 \\ .4 &.6\end{bmatrix}$
			&$\begin{bmatrix} .3 &.7 \\ .6 &.4\end{bmatrix}$\\
		\midrule
		$\Pi \ | \ x=1$
			&$\begin{bmatrix} 1 &0 \\ 0 &1\end{bmatrix}$
			&$\begin{bmatrix} 0 &1 \\ 1 &0\end{bmatrix}$
			&$\begin{bmatrix} .6 &.4 \\ .3 &.7\end{bmatrix}$
			&$\begin{bmatrix} .4 &.6 \\ .7 &.3\end{bmatrix}$\\
		\bottomrule
  	\end{tabular}
	\caption{CB Noncompliance Model Configurations}
	\label{tab:env_nc}
\end{table}
\begin{table*}[ht]
  \centering
  \small
  \begin{tabular}{ccccccccccccc}
    \toprule
	&\multicolumn{3}{c}{Environment 1}
	&\multicolumn{3}{c}{Environment 2}
	&\multicolumn{3}{c}{Environment 3}
	&\multicolumn{3}{c}{Environment 4}\\
	\cmidrule(r){2-4} \cmidrule(r){5-7} \cmidrule(r){8-10} \cmidrule(r){11-13}
    Algorithm	& Q-50			& Mean			& STD.
    				& Q-50			& Mean			& STD.
    				& Q-50			& Mean			& STD.
    				& Q-50			& Mean			& STD.\\
    \midrule
    TS			& 8.95			& 10.18			& 4.07
    				& 10.55			& 11.23			& 4.28
    				& 21.15			& 24.58			& 21.24
    				& 18.77			& 19.53			& 10.44\\
    \midrule
    TS-Check		& 9.55			& 11.47			& 6.21
    				& 397.3			& 398.5			& 10.71
    				&\textbf{4.35}	&\textbf{4.62}	&\textbf{1.98}
    				& 233.7			& 233.6			& 2.69\\
	TS-Obs		&\textbf{9.35}	&\textbf{10.49}	&\textbf{3.98}
				&\textbf{10.35}	&\textbf{11.92}	&\textbf{7.26}
    				& 5.09			& 5.56			& 2.49
    				&\textbf{5.07}	&\textbf{5.68}	&\textbf{2.13}\\
    \midrule
    TS-Lat-0		&\textbf{4.70}	&\textbf{5.49}	& 2.57
    				&\textbf{5.85}	&\textbf{6.24}	& 2.91
    				&\textbf{6.18}	& 32.39			& 58.58
    				& 8.58			& 28.36			& 39.67\\
    TS-Lat-40	& 16.45			& 16.63			&\textbf{2.38}
    				& 17.00			& 17.16			&\textbf{2.21}
    				& 6.48			&\textbf{14.18}	&\textbf{25.63}
    				&\textbf{7.32}	&\textbf{18.85}	&\textbf{32.75}\\
    \bottomrule
  \end{tabular}
  \caption{Results for Noncompliant CB Environments}
  \label{tab:nc_results}
\end{table*}

We present empirical regret (equation \ref{eq:regret_nc}) measures of these simulations in table \ref{tab:nc_results}. With horizontal lines, we separate the models according to their assumptions: assumed compliance (TS), observed compliance (TS-Check \& TS-Obs), and latent compliance (TS-Lat-0 \& TS-Lat-40). For the latter two categories we bold top performances. We see that TS, TS-Check, and TS-Obs all behave similarily to \cref{sec:sim:mab}'s results. In the case of deterministic compliance (environments 1 \& 2), TS-Lat-0 outperforms TS. Here also, TS-Lat-40's incurred regret offset (from uniform exploration) places it behind TS. In the case of stochastic  compliance (environments 3 \& 4), TS-Lat-0 outperforms TS in the median, but not the mean due to to its long-tail failures. Here also, TS-Lat-40's soft-start mechanics allow it to outperform TS across the board by seemingly reducing performance variance and its corresponding long-tail failures.

Our theory regarding TS-Lat-0's long-tail failures in environments 3 \& 4, is that Variational Inference minimizes $\DKL{q}{p}$. This direction for KL-Divergence is notoriously mode-seeking. For regions on the support where $q$ densities are low, there is a suppressed weighting of the corresponding divergence term $\log\frac{q}{p}$. Consequently, the posterior approximation $q$ can seriously underestimate the entropy and miss modes of $p$. This underestimation can occur when the first samples of $(r_t,a_t,z_t)$ go against their expectations. In this scenario, the $q$ approximation may incorrectly underestimate the true variance, which consequently prohibits future exploration and prematurely commits the bandit to a sub-optimal arm. Conversely, if enough early samples go with their expectations, then TS-Lat can advantageously commit to the optimal arm earlier than the other algorithms as evidenced in environments 1 \& 2.

\section{Experiments}
\label{sec:exp}
\begin{figure}[ht]
\centering
\includegraphics[scale=0.2]{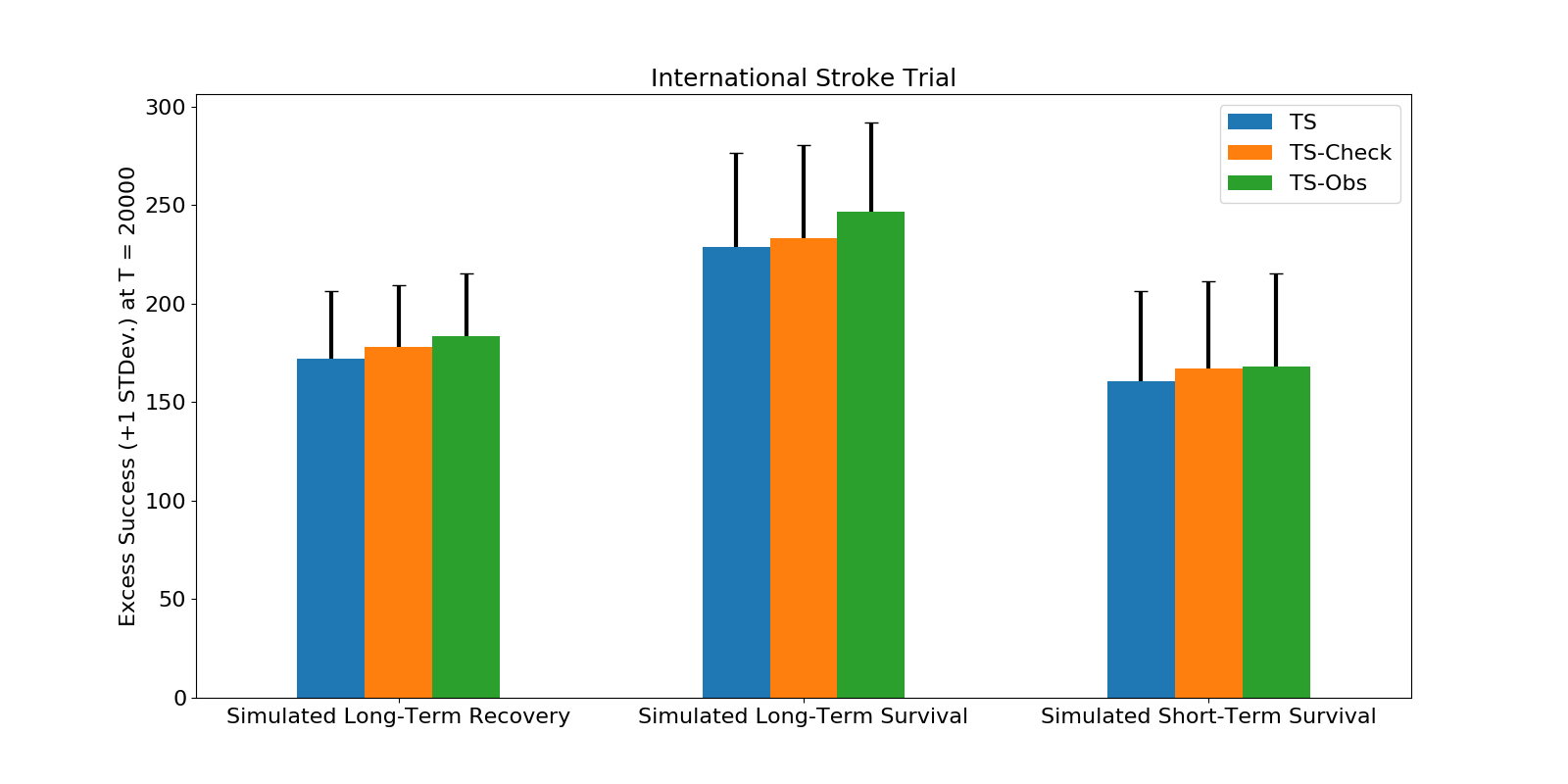}
\caption{Excess Stroke Trial Outcomes}
\label{fig:stroke}
\end{figure}

The International Stroke Trial (IST) \citep{ref:SandercockIST} was a randomized clinical trial that studied the effects of Aspirin and Heparin on stroke victims. The study's action space is $\mathcal{A} \times \mathcal{H}$ where
\begin{equation*}
\begin{aligned}
	\mathcal{A} &= \{\text{no aspirin}, \text{aspirin}\}\\
	\mathcal{H} &= \{\text{no heparin}, \text{lo dose heparin}, \text{hi dose heparin}\}.
\end{aligned}
\end{equation*}
Amazingly, the IST data has compliance outcomes for all 19,435 admitted patients. For each patient, the trial recorded short-term survival (STS--discharged alive in 14 days), long-term survival (LTS--alive in 6 months), and long-term recovery (LTR--fully recovered at 6 months) outcomes. We consider each of these as a Bernoulli reward variable. To conduct our analysis, we empirically computed the compliance and reward parameters across all subjects:
\begin{equation*}
\begin{aligned}
	\Pi &= \begin{bmatrix}
		.980 &.002 &.002 &.014 &.001 &.001\\
		.000 &.975 &.009 &.000 &.014 &.002\\
		.000 &.005 &.983 &.000 &.000 &.012\\
		.068 &.001 &.001 &.928 &.000 &.001\\
		.000 &.102 &.001 &.000 &.882 &.015\\
		.000 &.001 &.082 &.000 &.004 &.914\\
	\end{bmatrix}\\
	\mu_\text{STS} &= [.886, .886, .888, .903, .896, .910]^T\\
	\mu_\text{LTS} &= [.760, .749, .747, .785, .775, .782]^T\\
	\mu_\text{LTR} &= [.181, .178, .181, .201, .208, .206]^T
\end{aligned}
\end{equation*}
For TS, TS-Check, and TS-Obs, we used these environmental parameters to simulate a 20,000 patient trial 500 times according to the noncompliance Bernoulli bandit model in \cref{sec:alg}. We present the number of \enquote{excess successes} for each reward class in figure \ref{fig:stroke}. An \enquote{excess success} is the expected number of additional successes (the difference in final cumulative regret) over a uniform exploration policy, which is often employed in clinical trials. For STS and LTS, an \enquote{excess success} is quite literally a human life. For LTR, an \enquote{excess success} is a full recovery.

With high levels of compliance, we do not expect TS-Check and TS-Obs to significantly outperform TS. From our simulations, TS, TS-Check, and TS-Obs all exhibit significant life saving potential over the trial's uniform exploration policy. However, in a scenario where human lives are on the line, we must look to the absolute top performer, which in all three cases appears to be TS-Obs.

\section{Future Work and Conclusion}
Theorem \ref{thm:nc_k} identifies the existence of helpful and harmful noncompliance in $(K>2)$-armed bandits. Assuming access to $p(\pi|a)$, one should be able to bound the probabilities of observing harmful versus helpful compliance. Assuming a uniform $p(\pi|a)$, we conjecture that harmful noncompliance is highly more probable. We also consider it an open problem to derive problem-dependent expected regret bounds for out TS-Obs algorithm to the rigor of \cite{ref:agrawal_mab_bound_2}.

In this article, we consider noncompliance in bandit problems. We present a probabilistic model that captures the stochastic mapping of an agent's chose action to implemented action. We outline several real-world scenarios where stochastic noncompliance can occur. Further, we proved that for 2-armed Bernoulli bandits existing expected regret bounds \citep{ref:agrawal_mab_bound_2} are a lower bound for expected regret in the presence of noncompliance. In contrast to previous works, we take a model-first approach in devising suitable algorithms for the noncompliant bandit. Following Bayesian principles, our model leads naturally to Thompson sampling solutions for both the observed and latent noncompliance scenarios. To the best of our knowledge, we are the first to consider and present solutions for the latent noncompliance problem. Finally, we empirically validate our proposed solutions through extensive simulations.

\subsubsection*{Acknowledgments}
Work supported in part by the grant NSF III-1526914 Small: Collaborative Research: Approximate Learning and Inference in Graphical Models.

\bibliographystyle{plainnat}
\bibliography{references}

\begin{thebibliography}{24}
\providecommand{\natexlab}[1]{#1}
\providecommand{\url}[1]{\texttt{#1}}
\expandafter\ifx\csname urlstyle\endcsname\relax
  \providecommand{\doi}[1]{doi: #1}\else
  \providecommand{\doi}{doi: \begingroup \urlstyle{rm}\Url}\fi

\bibitem[Agrawal and Goyal(2012)]{ref:agrawal_mab_bound_1}
Shipra Agrawal and Navin Goyal.
\newblock Analysis of thompson sampling for the multi-armed bandit problem.
\newblock In \emph{{COLT} 2012 - The 25th Annual Conference on Learning Theory,
  June 25-27, 2012, Edinburgh, Scotland}, pages 39.1--39.26, 2012.

\bibitem[Agrawal and Goyal(2013{\natexlab{a}})]{ref:agrawal_cb_bound}
Shipra Agrawal and Navin Goyal.
\newblock Thompson sampling for contextual bandits with linear payoffs.
\newblock In \emph{Proceedings of the 30th International Conference on Machine
  Learning, {ICML} 2013, Atlanta, GA, USA, 16-21 June 2013}, pages 127--135,
  2013{\natexlab{a}}.

\bibitem[Agrawal and Goyal(2013{\natexlab{b}})]{ref:agrawal_mab_bound_2}
Shipra Agrawal and Navin Goyal.
\newblock Further optimal regret bounds for thompson sampling.
\newblock In \emph{Proceedings of the Sixteenth International Conference on
  Artificial Intelligence and Statistics, {AISTATS} 2013, Scottsdale, AZ, USA,
  April 29 - May 1, 2013}, pages 99--107, 2013{\natexlab{b}}.

\bibitem[Arora et~al.(2018)Arora, Cohen, and Hazan]{ref:op_2}
Sanjeev Arora, Nadav Cohen, and Elad Hazan.
\newblock On the optimization of deep networks: Implicit acceleration by
  overparameterization.
\newblock In Jennifer Dy and Andreas Krause, editors, \emph{Proceedings of the
  35th International Conference on Machine Learning}, volume~80 of
  \emph{Proceedings of Machine Learning Research}, pages 244--253,
  Stockholmsmässan, Stockholm Sweden, 10--15 Jul 2018. PMLR.

\bibitem[Bareinboim et~al.(2015)Bareinboim, Forney, and Pearl]{ref:cf_bandits}
Elias Bareinboim, Andrew Forney, and Judea Pearl.
\newblock Bandits with unobserved confounders: A causal approach.
\newblock In C.~Cortes, N.~D. Lawrence, D.~D. Lee, M.~Sugiyama, and R.~Garnett,
  editors, \emph{Advances in Neural Information Processing Systems 28}, pages
  1342--1350. Curran Associates, Inc., 2015.

\bibitem[Chapelle and Li(2011)]{ref:chapelle_ts_empirical}
Olivier Chapelle and Lihong Li.
\newblock An empirical evaluation of thompson sampling.
\newblock In \emph{Advances in Neural Information Processing Systems 24: 25th
  Annual Conference on Neural Information Processing Systems 2011. Proceedings
  of a meeting held 12-14 December 2011, Granada, Spain.}, pages 2249--2257,
  2011.

\bibitem[Du and Lee(2018)]{ref:op_3}
Simon Du and Jason Lee.
\newblock On the power of over-parametrization in neural networks with
  quadratic activation.
\newblock In Jennifer Dy and Andreas Krause, editors, \emph{Proceedings of the
  35th International Conference on Machine Learning}, volume~80 of
  \emph{Proceedings of Machine Learning Research}, pages 1329--1338,
  Stockholmsmässan, Stockholm Sweden, 10--15 Jul 2018. PMLR.

\bibitem[Honda and Takemura(2014)]{ref:priors}
Junya Honda and Akimichi Takemura.
\newblock Optimality of thompson sampling for gaussian bandits depends on
  priors.
\newblock In \emph{{AISTATS}}, volume~33 of \emph{{JMLR} Workshop and
  Conference Proceedings}, pages 375--383. JMLR.org, 2014.

\bibitem[Jordan et~al.(1999)Jordan, Ghahramani, Jaakkola, and Saul]{ref:vi}
Michael~I. Jordan, Zoubin Ghahramani, Tommi~S. Jaakkola, and Lawrence~K. Saul.
\newblock An introduction to variational methods for graphical models.
\newblock \emph{Machine Learning}, 37\penalty0 (2):\penalty0 183--233, 1999.

\bibitem[Kallus(2018)]{ref:iab}
Nathan Kallus.
\newblock Instrument-armed bandits.
\newblock In \emph{Algorithmic Learning Theory, {ALT} 2018, 7-9 April 2018,
  Lanzarote, Canary Islands, Spain}, pages 529--546, 2018.

\bibitem[Korda et~al.(2013)Korda, Kaufmann, and Munos]{ref:korda_optimality}
Nathaniel Korda, Emilie Kaufmann, and R{\'{e}}mi Munos.
\newblock Thompson sampling for 1-dimensional exponential family bandits.
\newblock In \emph{Advances in Neural Information Processing Systems 26: 27th
  Annual Conference on Neural Information Processing Systems 2013. Proceedings
  of a meeting held December 5-8, 2013, Lake Tahoe, Nevada, United States.},
  pages 1448--1456, 2013.

\bibitem[Li et~al.(2010{\natexlab{a}})Li, Chu, Langford, and
  Schapire]{ref:li_CB}
Lihong Li, Wei Chu, John Langford, and Robert~E. Schapire.
\newblock A contextual-bandit approach to personalized news article
  recommendation.
\newblock \emph{CoRR}, abs/1003.0146, 2010{\natexlab{a}}.

\bibitem[Li et~al.(2010{\natexlab{b}})Li, Chu, Langford, and
  Schapire]{ref:li_ts_content}
Lihong Li, Wei Chu, John Langford, and Robert~E. Schapire.
\newblock A contextual-bandit approach to personalized news article
  recommendation.
\newblock \emph{CoRR}, abs/1003.0146, 2010{\natexlab{b}}.

\bibitem[Liao and Couillet(2018)]{ref:op_1}
Zhenyu Liao and Romain Couillet.
\newblock The dynamics of learning: A random matrix approach.
\newblock In Jennifer Dy and Andreas Krause, editors, \emph{Proceedings of the
  35th International Conference on Machine Learning}, volume~80 of
  \emph{Proceedings of Machine Learning Research}, pages 3072--3081,
  Stockholmsmässan, Stockholm Sweden, 10--15 Jul 2018. PMLR.

\bibitem[Penna et~al.(2016)Penna, Reid, and Balduzzi]{ref:comp_aware}
Nicol{\'{a}}s~Della Penna, Mark~D. Reid, and David Balduzzi.
\newblock Compliance-aware bandits.
\newblock \emph{CoRR}, abs/1602.02852, 2016.

\bibitem[Russo and Roy(2014)]{ref:russo_l2o}
Daniel Russo and Benjamin~Van Roy.
\newblock Learning to optimize via posterior sampling.
\newblock \emph{Math. Oper. Res.}, 39\penalty0 (4):\penalty0 1221--1243, 2014.
\newblock \doi{10.1287/moor.2014.0650}.

\bibitem[Russo and Roy(2016)]{ref:russo_info_thry}
Daniel Russo and Benjamin~Van Roy.
\newblock An information-theoretic analysis of thompson sampling.
\newblock \emph{Journal of Machine Learning Research}, 17:\penalty0
  68:1--68:30, 2016.

\bibitem[S.~S.~Dragomir(2000)]{ref:kl_bound}
J.~Sunde S.~S.~Dragomir, M. L.~Scholz.
\newblock Some upper bounds for relative entropy and applications.
\newblock \emph{Computers and Mathematics with Applications}, 2000.

\bibitem[Sabat{\'{e}}(2000)]{ref:nc_who}
Eduardo Sabat{\'{e}}.
\newblock Adherence to long-term therapies: Evidence for action.
\newblock \emph{World Health Organization}, 2000.

\bibitem[Sandercock et~al.(2011)Sandercock, Niewada, Cz{\l}onkowska, and {the
  International Stroke Trial Collaborative Group}]{ref:SandercockIST}
Peter~AG Sandercock, Maciej Niewada, Anna Cz{\l}onkowska, and {the
  International Stroke Trial Collaborative Group}.
\newblock The international stroke trial database.
\newblock \emph{Trials}, 12\penalty0 (1):\penalty0 101, Apr 2011.
\newblock ISSN 1745-6215.
\newblock \doi{10.1186/1745-6215-12-101}.

\bibitem[Sutton and Barto(1998)]{ref:sutton_barto}
Richard~S. Sutton and Andrew~G. Barto.
\newblock \emph{Reinforcement learning - an introduction}.
\newblock Adaptive computation and machine learning. {MIT} Press, 1998.
\newblock ISBN 0262193981.

\bibitem[Thompson(1933)]{ref:ts}
W.~R. Thompson.
\newblock On the likelihood that one unknown probability exceeds another in
  view of the evidence of two samples.
\newblock \emph{Biometrika}, 25:\penalty0 285--294, 1933.

\bibitem[Urteaga and Wiggins(2018)]{ref:urteaga_vi_ts}
I{\~{n}}igo Urteaga and Chris Wiggins.
\newblock Variational inference for the multi-armed contextual bandit.
\newblock In \emph{International Conference on Artificial Intelligence and
  Statistics, {AISTATS} 2018, 9-11 April 2018, Playa Blanca, Lanzarote, Canary
  Islands, Spain}, pages 698--706, 2018.

\bibitem[Wainwright and Jordan(2008)]{ref:vi_exp}
Martin~J. Wainwright and Michael~I. Jordan.
\newblock Graphical models, exponential families, and variational inference.
\newblock \emph{Foundations and Trends in Machine Learning}, 1\penalty0
  (1-2):\penalty0 1--305, 2008.

\end{thebibliography}

\newpage
\onecolumn
\section{Appendix}
\label{sec:appendix}

\begin{theorem}[An upper bound for KL-Divergence \citep{ref:kl_bound}]
\label{thm:dkl_bnd}
Let $p(x), q(x)>0, x \in \X$ be two probability mass functions.  Then
\begin{equation}
\begin{aligned}
	\DKL{p}{q} \leq \sum_{x\in\mathcal{X}} \frac{p^2(x)}{q(x)} - 1
\end{aligned}
\end{equation}
\end{theorem}

\subsection{Proof of Lemma \ref{lemma:grad}}
\label{sec:appendix:lemma1_proof}
To prove $\frac{\partial}{\partial \mu_1}f(\mu_1,\mu_i) \leq 0 \ \forall \ \mu \in \text{dom } f$, we first derive an equivalent condition:
\begin{equation}
\begin{aligned}
	\frac{\partial}{\partial \mu_1}f(\mu_1,\mu_i) \leq 0
		&\iff \frac{(\mu_1 - \mu_i)\Big(\frac{\mu_i}{\mu_1}-\frac{1-\mu_i}{1-\mu_1}\Big)}
			{(\DKL{\mu_i}{\mu_1})^2} + \frac{1}{\DKL{\mu_i}{\mu_1}} \leq 0\\
		&\iff (\mu_1 - \mu_i)\Big(\frac{\mu_i}{\mu_1}-\frac{1-\mu_i}{1-\mu_1}\Big)
			+ \DKL{\mu_i}{\mu_1} \leq 0\\
		&\iff \DKL{\mu_i}{\mu_1} \leq 
			(\mu_i - \mu_1)\Big(\frac{\mu_i}{\mu_1}-\frac{1-\mu_i}{1-\mu_1}\Big)\\
\end{aligned}
\end{equation}
Beginning with theorem \ref{thm:dkl_bnd}, we prove the the last of the above equivalent conditions:
\begin{equation}
\begin{aligned}
	\DKL{\mu_i}{\mu_1}
		&\leq \frac{\mu_i^2}{\mu_1} + \frac{(1-\mu_i)^2}{(1-\mu_1)} - 1\\
		&= \frac{\mu_i^2(1-\mu_1) + \mu_1(1-\mu_i)^2}{\mu_1(1-\mu_1)} - 1\\
		&= \frac{\mu_i^2-\mu_1\mu_i^2 + \mu_1(1-2\mu_i+\mu_i^2)}{\mu_1(1-\mu_1)} - 1\\
		&= \frac{\mu_i^2-\mu_1\mu_i^2 + \mu_1 - 2\mu_1\mu_i+\mu_1\mu_i^2}{\mu_1(1-\mu_1)} - 1\\
		&= \frac{\mu_i^2 - 2\mu_1\mu_i + \mu_1^2 + \mu_1 - \mu_1^2}{\mu_1(1-\mu_1)} - 1\\
		&= \frac{(\mu_i-\mu_1)^2 + \mu_1(1-\mu_1)}{\mu_1(1-\mu_1)} - 1\\
		&= \frac{(\mu_i-\mu_1)^2}{\mu_1(1-\mu_1)}\\
		&= (\mu_i-\mu_1)\Big(\frac{\mu_i-\mu_1\mu_i-\mu_1+\mu_1\mu_i}{(1-\mu_1)\mu_1}\Big)\\
		&= (\mu_i-\mu_1)\Big(\frac{\mu_i(1-\mu_1)}{\mu_1(1-\mu_1)}-\frac{(1-\mu_i)\mu_1}{(1-\mu_1)\mu_1}\Big)\\
		&= (\mu_i-\mu_1)\Big(\frac{\mu_i}{\mu_1}-\frac{1-\mu_i}{1-\mu_1}\Big)\\
\end{aligned}
\end{equation}
To prove $\frac{\partial}{\partial \mu_i}f(\mu_1,\mu_i) \geq 0 \ \forall \ \mu \in \text{dom } f$, observe:
\begin{equation}
\begin{aligned}
	\frac{\partial}{\partial \mu_i}f(\mu_1,\mu_i)
		&= \frac{(\mu_1 - \mu_i)\Big(-\log\frac{\mu_i}{\mu_1} + \log\frac{1-\mu_i}{1-\mu_1}\Big)}
			{\Big(\mu_i\log\frac{\mu_i}{\mu_1}+(1-\mu_i)\log\frac{1-\mu_i}{1-\mu_1}\Big)^2}
		 - \frac{1}
		 	{\Big(\mu_i\log\frac{\mu_i}{\mu_1}+(1-\mu_i)\log\frac{1-\mu_i}{1-\mu_1}\Big)}\\
		&= \frac{
		 	-\mu_1\log\frac{\mu_i}{\mu_1} + \mu_1\log\frac{1-\mu_i}{1-\mu_1}
		 	+\mu_i\log\frac{\mu_i}{\mu_1} - \mu_i\log\frac{1-\mu_i}{1-\mu_1}}
		 	{(\DKL{\mu_i}{\mu_1})^2} - \frac{1}{\DKL{\mu_i}{\mu_1}}\\
		&= \frac{
		 	-\mu_1\log\frac{\mu_i}{\mu_1} + \mu_1\log\frac{1-\mu_i}{1-\mu_1}
		 	+\mu_i\log\frac{\mu_i}{\mu_1} - \mu_i\log\frac{1-\mu_i}{1-\mu_1}}
		 	{(\DKL{\mu_i}{\mu_1})^2}
		  - \frac{\mu_i\log\frac{\mu_i}{\mu_1}+(1-\mu_i)\log\frac{1-\mu_i}{1-\mu_1}}
		  	{(\DKL{\mu_i}{\mu_1})^2}\\
		&= \frac{
		 	-\mu_1\log\frac{\mu_i}{\mu_1} + \mu_1\log\frac{1-\mu_i}{1-\mu_1}
		 	- \log\frac{1-\mu_i}{1-\mu_1}}
		  	{(\DKL{\mu_i}{\mu_1})^2}\\
		&= \frac{
		 	\mu_1\log\frac{\mu_1}{\mu_i} + (1 - \mu_1)\log\frac{1-\mu_1}{1-\mu_i}}
		  	{(\DKL{\mu_i}{\mu_1})^2}\\
		&= \frac{\DKL{\mu_1}{\mu_i}}
		  	{(\DKL{\mu_i}{\mu_1})^2}\\
		&\geq 0
\end{aligned}
\end{equation}

\subsection{Latent Noncompliance Posterior Derivation}
\label{sec:vi_deriv}
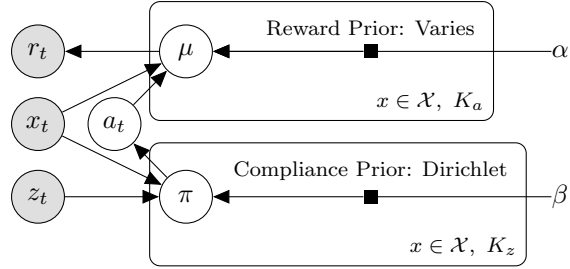
\begin{figure}[H]
\centering
\begin{tikzpicture}

	\node[obs] (x) {$x_t$};
	\node[latent, right=0.3cm of x] (a) {$a_t$};

	\node[obs, above=0.25cm of x] (r) {$r_t$};
	\node[latent, right=1.25cm of r] (mu) {$\mu$};	
	\node[const, right=4.5cm of mu] (alpha) {$\alpha$};
	
	\node[obs, below=0.25cm of x] (z) {$z_t$};
	\node[latent, right=1.25cm of z] (pi) {$\pi$};
	\node[const, right=4.5cm of pi] (beta) {$\beta$};
	
	\factor[right=2cm of mu] {mu-f} {above:Reward Prior: Varies} {alpha} {mu}; 	
	\factor[right=2cm of pi] {pi-f} {above:Compliance Prior: Dirichlet} {beta} {pi}; 
			
	\edge {x} {mu, pi};
	\edge {mu-f} {mu};
	\edge {a} {mu};
	\edge {mu} {r};
	\edge {pi} {a};
	\edge {z} {pi};
	
	\plate {p-mu} {(mu)(mu-f)(mu-f-caption)} {$x \in \X, \ K_a$};
	\plate {p-pi} {(pi)(pi-f)(pi-f-caption)} {$x \in \X, \ K_z$};

\end{tikzpicture}
\caption{Contextual Bandit with Observed NC}
\label{fig:gm_lat}
\end{figure}

The graphical model for an environment with latent noncompliance appears in figure \ref{fig:gm_lat}.
In \cref{sec:alg}, we consider a Bernoulli MAB such that we can disregard the a priori context variable $x_t$. We first compute the joint likelihood and log joint likelihood that we will need for Variational Inference:
\begin{equation*}
\begin{aligned}
	p(r, a, \mu, \pi)
		=&	p(r|\mu,a)p(a|\pi,z)p(\mu|a)p(\pi|z)\\
		=&	\Bigg[\prod_{i=1}^N \prod_{j=1}^K p(r_i|\mu_j)^{\ictr[a_i = j]}\Bigg]
			\Bigg[\prod_{i=1}^N \prod_{j=1}^K \prod_{k=1}^K p(a_i = j | \pi_k)^{\ictr[z_i = k]}\Bigg]
			\Bigg[\prod_{j=1}^K p(\mu_j)\Bigg]
			\Bigg[\prod_{k=1}^K p(\pi_k)\Bigg] \\
	\ln p(r, a, \mu, \pi)
		=&	\Bigg[\sum_{i=1}^N \sum_{j=1}^K \ictr[a_i = j] \ln p(r_i|\mu_j)\Bigg] +
			\Bigg[\sum_{i=1}^N \sum_{j=1}^K \sum_{k=1}^K \ictr[z_i = k] \ln p(a_i = j | \pi_k)\Bigg] +\\
		&	\Bigg[\sum_{j=1}^K \ln p(\mu_j)\Bigg] + 
			\Bigg[\sum_{k=1}^K \ln p(\pi_k)\Bigg] \\
\end{aligned}
\end{equation*}
Through Variational Inference, we seek to approximate the posterior as follows:
\begin{equation}
\begin{aligned}
	p(a, \mu, \pi| z, r) \approx q(a, \mu, \pi) =
		\Bigg[\prod_{i=1}^N q(a_i)\Bigg]
		\Bigg[\prod_{j=1}^K q(\mu_j)\Bigg]
		\Bigg[\prod_{k=1}^K q(\pi_k)\Bigg]
\end{aligned}
\end{equation}
Since our model and prior distributions are from the Conjugate-Exponential family, we know at the onset:
\begin{equation}
\begin{aligned}
	q(a_i) &= \Cat{\phi_i(1), \hdots, \phi_i(K)}\\
	q(\mu_j) &= \Bet{\alpha'_{1,j}, \alpha'_{0,j}}\\
	q(\pi_k) &= \Dir{\beta'_{k,1}, \hdots, \beta'_{k,K}}\\
\end{aligned}
\end{equation}
Additionally, we know that the variational parameters for these distributions can be found by normalizing the exponential expectation of the log joint probability. If $y$ is the variable for which we seek variational parameter updates, then we take the expectation with respect to all variational distributions except the one for $y$ as shown: 
\begin{equation}
\begin{aligned}
	q(y) \varpropto& \exp\Bigg(\E{q(\{a, \mu, \pi\} \setminus y)} \ln p(r, a, \mu, \pi) \Bigg)\\\\
\end{aligned}
\end{equation}
Using this strategy, we derive parameter updates for $q(\mu_j)$ and therefore all $\mu_j$:		
\begin{equation}
\begin{aligned}
	q(\mu_j)
		\varpropto& 
			\exp\Bigg(\E{q(a)q(\pi)}
				\Bigg[\sum_{i=1}^N \sum_{j=1}^K \ictr[a_i = j] \ln p(r_i|\mu_j)\Bigg]
			\Bigg)
			\exp\Bigg(\E{q(a)q(\pi)}\Bigg[\ln p(\mu_j)\Bigg]\Bigg)\\
		\varpropto& 
			\exp\Bigg(\E{q(a)q(\pi)}\Bigg[\sum_{i=1}^N \ictr[a_i = j] \ln p(r_i|\mu_j)\Bigg]\Bigg)
			\exp\Bigg(\E{q(a)q(\pi)}\Bigg[\ln p(\mu_j)\Bigg]\Bigg)\\
		\varpropto& 
			\exp\Bigg(\E{q(a)}\Bigg[\sum_{i=1}^N \ictr[a_i = j] \ln\Big( (\mu_j)^{(r_i)} \cdot(1-\mu_j)^{(1 -r_i)} \Big)\Bigg]\Bigg)
			\exp\Bigg(\ln p(\mu_j)\Bigg)\\
		=& 
			\exp\Bigg(\sum_{i=1}^N \E{q(a)}\Big[\ictr[a_i = j]\Big] \ln\Big( (\mu_j)^{(r_i)} \cdot(1-\mu_j)^{(1 -r_i)} \Big)\Bigg)
			(\mu_j)^{(\alpha_1-1)}(1 - \mu_j)^{(\alpha_0-1)}\\
		=& 
			\Bigg[\prod_{i=1}^N \exp\Bigg(\phi_i(j) \ln\Big( (\mu_j)^{(r_i)} \cdot(1-\mu_j)^{(1 -r_i)} \Big)\Bigg)\Bigg]
			(\mu_j)^{(\alpha_1-1)}(1 - \mu_j)^{(\alpha_0-1)}\\
		=& 
			\Bigg[\prod_{i=1}^N \exp\Bigg(\ln\Big( (\mu_j)^{r_i\phi_i(j)} (1-\mu_j)^{(1 -r_i)\phi_i(j)} \Big)\Bigg)\Bigg]
			(\mu_j)^{(\alpha_1-1)}(1 - \mu_j)^{(\alpha_0-1)}\\
		=& 
			\Bigg[\prod_{i=1}^N (\mu_j)^{r_i\phi_i(j)} (1-\mu_j)^{(1 -r_i)\phi_i(j)} \Bigg]
			(\mu_j)^{(\alpha_1-1)}(1 - \mu_j)^{(\alpha_0-1)}\\
		=& 
			(\mu_j)^{\Big(\sum\limits_{i=1}^N r_i\phi_i(j)\Big)} (1-\mu_j)^{\Big(\sum\limits_{i=1}^N (1 -r_i)\phi_i(j)\Big)}
			(\mu_j)^{(\alpha_1-1)}(1 - \mu_j)^{(\alpha_0-1)}\\
		=& 
			(\mu_j)^{\Big(\alpha_1 + \Big[\sum\limits_{i=1}^N r_i\phi_i(j)\Big] - 1 \Big)}
			(1-\mu_j)^{\Big(\alpha_0 + \Big[\sum\limits_{i=1}^N (1 -r_i)\phi_i(j)\Big] - 1\Big)}\\
	\alpha'_{1,j}
		=& \alpha_1 + \sum\limits_{i=1}^N r_i\phi_i(j)\\
	\alpha'_{0,j}
		=& \alpha_0 + \sum\limits_{i=1}^N (1 -r_i)\phi_i(j)\\
\end{aligned}
\end{equation}	
We now do the same for $q(\pi_k)$, which again results in solutions for all $\pi_k$:
\begin{equation}
\begin{aligned}
	q(\pi_k)
		\varpropto& 
			\exp\Bigg(\E{q(a)q(\mu)}
				\Bigg[\sum_{i=1}^N \sum_{j=1}^K \sum_{k=1}^K \ictr[z_i = k] \ln p(a_i = j | \pi_k)\Bigg]
			\Bigg)
			\exp\Bigg(\E{q(a)q(\mu)}\Bigg[\ln p(\pi_k)\Bigg]\Bigg)\\
		\varpropto& 
			\exp\Bigg(\E{q(a)q(\mu)}\Bigg[\sum_{i=1}^N \sum_{j=1}^K \ictr[z_i = k] \ln p(a_i = j | \pi_k)\Bigg]\Bigg)
			\exp\Bigg(\E{q(a)q(\mu)}\Bigg[\ln p(\pi_k)\Bigg]\Bigg)\\
		\varpropto&
			\exp\Bigg(\E{q(a)}\Bigg[\sum_{i=1}^N \sum_{j=1}^K \ictr[z_i = k] \ln (\pi_{k,j}^{\ictr[a_i = j]}) \Bigg]\Bigg)
			\exp\Bigg(\ln \prod_{j=1}^K \pi_{k,j}^{(\beta - 1)}\Bigg)\\
		=&
			\exp\Bigg(\sum_{i=1}^N \sum_{j=1}^K \ictr[z_i = k] \E{q(a)q(\mu)}\Big[\ictr[a_i = j] \Big] \ln (\pi_{k,j}) \Bigg)
			\Bigg[\prod_{j=1}^K \pi_{k,j}^{(\beta - 1)}\Bigg]\\
		=&
			\Bigg[\prod_{i=1}^N \prod_{j=1}^K\exp\Bigg( \ictr[z_i = k] \cdot \phi_i(j) \cdot \ln (\pi_{k,j}) \Bigg)\Bigg]
			\Bigg[\prod_{j=1}^K \pi_{k,j}^{(\beta - 1)}\Bigg]\\
		=&
			\Bigg[\prod_{i=1}^N \prod_{j=1}^K \pi_{k,j}^{(\ictr[z_i = k] \cdot \phi_i(j))}\Bigg]
			\Bigg[\prod_{j=1}^K \pi_{k,j}^{(\beta - 1)}\Bigg]\\
		=&
			\Bigg[\prod_{j=1}^K \pi_{k,j}^{\Big(\sum\limits_{i=1}^N \ictr[z_i = k] \cdot \phi_i(j)\Big)}\Bigg]
			\Bigg[\prod_{j=1}^K \pi_{k,j}^{(\beta - 1)}\Bigg]\\
		=&
			\Bigg[\prod_{j=1}^K \pi_{k,j}^{\Big(\beta + \Big[\sum\limits_{i=1}^N \ictr[z_i = k] \cdot \phi_i(j)\Big] - 1\Big)}\Bigg]\\
	\beta'_{k,j}
		=& \beta + \sum_{i=1}^N \ictr[z_i = k] \cdot \phi_i(j)\\
\end{aligned}
\end{equation}	
Finally, we derive the parameters for $q(a_i = j)$ and therefore all possible implemented actions across time:
{\small \begin{equation}
\begin{aligned}
	q(a_i = j)
		\varpropto& \exp\Bigg(\E{q(a_i \neq j)q(\mu)q(\pi)} \Bigg[
			\sum_{i=1}^N \sum_{j=1}^K \ictr[a_i = j] \ln p(r_i|\mu_j) +
			\sum_{i=1}^N \sum_{j=1}^K \sum_{k=1}^K \ictr[z_i = k] \ln p(a_i = j | \pi_k)\Bigg]
		\Bigg)\\
		\varpropto& \exp\Bigg(\E{q(\mu)q(\pi)} \Bigg[
			\ln p(r_i|\mu_j) + 
			\ln p(a_i = j | \pi_{z_i})\Bigg]
		\Bigg)\\
		=& \exp\Bigg(
			\E{q(\mu_j)} \Bigg[\ln\Big( (\mu_j)^{r_i} (1-\mu_j)^{(1 -r_i)} \Big)\Bigg] + 
			\E{q(\pi_{z_i})} \Bigg[\ln \pi_{z_i,j}\Bigg]
		\Bigg)\\
		=& \exp\Bigg(
			r_i \E{q(\mu_j)} \Big[\ln \mu_j\Big] +
			(1-r_i) \E{q(\mu_j)} \Big[\ln (1 - \mu_j)\Big] + 
			\E{q(\pi_{z_i})} \Big[\ln \pi_{z_i,j}\Big]
		\Bigg)\\
		\varpropto& \frac{\exp\Bigg(
			r_i \E{q(\mu_j)} \Big[\ln \mu_j\Big] +
			(1-r_i) \E{q(\mu_j)} \Big[\ln (1 - \mu_j)\Big] + 
			\E{q(\pi_{z_i})} \Big[\ln \pi_{z_i,j}\Big]
		\Bigg)}{\sum\limits_{k=1}^K \exp\Bigg(
			r_i \E{q(\mu_k)} \Big[\ln \mu_k\Big] +
			(1-r_i) \E{q(\mu_k)} \Big[\ln (1 - \mu_k)\Big] + 
			\E{q(\pi_{z_i})} \Big[\ln \pi_{z_i,k}\Big]
		\Bigg)}\\
		=& \frac{\exp\Bigg(
			r_i \Big(\psi(\alpha'_{1,j}) - \psi(\alpha'_{1,j} + \alpha'_{0,j})\Big)+
			(1-r_i) \Big(\psi(\alpha'_{0,j}) - \psi(\alpha'_{1,j} + \alpha'_{0,j})\Big) +
			\psi(\beta'_{z_i,j}) - \psi\Big(\sum_{l=1}^K \beta'_{z_i,l}\Big)
		\Bigg)}{\sum\limits_{k=1}^K \exp\Bigg(
			r_i \Big(\psi(\alpha'_{k,1}) - \psi(\alpha'_{k,1} + \alpha'_{k,0})\Big)+
			(1-r_i) \Big(\psi(\alpha'_{k,0}) - \psi(\alpha'_{k,1} + \alpha'_{k,0})\Big) +
			\psi(\beta'_{z_i,k}) - \psi\Big(\sum_{l=1}^K \beta'_{z_i,l}\Big)
		\Bigg)}\\
		=& \phi_i(j)\\
\end{aligned}
\end{equation}}
To assess convergence, we wish to derive the variational objective, the Evidence Lower Bound (ELBO), as a function of our variational parameters:
\begin{equation}
\begin{aligned}
	\mathcal{L}(\phi, \alpha', \beta')
		=&
			\E{q(a)q(\mu)q(\pi)} \Big[ \ln p(r, a, \mu, \pi) \Big] -
			\E{q(a)} \Big[ \ln q(a) \Big] -
			\E{q(\mu)} \Big[ \ln q(\mu) \Big] -
			\E{q(\pi)} \Big[ \ln q(\pi) \Big]\\
		=&
		 	\E{q(a)q(\mu)q(\pi)} \Bigg[\sum_{i=1}^N \sum_{j=1}^K \ictr[a_i = j] \ln p(r_i|\mu_j)\Bigg] +  
			\E{q(a)q(\mu)q(\pi)} \Bigg[\sum_{i=1}^N \sum_{j=1}^K \sum_{k=1}^K \ictr[z_i = k] \ln p(a_i = j | \pi_k)\Bigg] + \\
		&	\E{q(a)q(\mu)q(\pi)} \Bigg[\sum_{j=1}^K \ln p(\mu_j)\Bigg] + 
			\E{q(a)q(\mu)q(\pi)} \Bigg[\sum_{k=1}^K \ln p(\pi_k)\Bigg] - \\
		&	\E{q(a)} \Bigg[ \sum_{i=1}^N \ln q(a_i) \Bigg] -
			\E{q(\mu)} \Bigg[ \sum_{j=1}^K \ln q(\mu_j) \Bigg] -
			\E{q(\pi)} \Bigg[ \sum_{k=1}^K \ln q(\pi_k) \Bigg]\\
\end{aligned}
\end{equation}
We now compute the seven expectations appearing in the ELBO. We note the last three are entropies.
First expectation:
\begin{equation}
\begin{aligned}
	\E{q(a)q(\mu)q(\pi)} \Bigg[&\sum_{i=1}^N \sum_{j=1}^K \ictr[a_i = j] \ln p(r_i|\mu_j)\Bigg]\\
		=&
			\sum_{i=1}^N \sum_{j=1}^K
			\E{q(a)q(\mu)q(\pi)} \Bigg[\ictr[a_i = j] \ln\Big( (\mu_j)^{r_i} (1-\mu_j)^{(1 -r_i)} \Big)\Bigg]\\
		=&
			\sum_{i=1}^N \sum_{j=1}^K
			\E{q(a)q(\mu)q(\pi)} \Bigg[\ictr[a_i = j] \Big( r_i \ln \mu_j + (1 -r_i) \ln (1-\mu_j) \Big)\Bigg]\\	
		=&
			\sum_{i=1}^N \sum_{j=1}^K
			\E{q(a)}\Big[\ictr[a_i = j]\Big] \Bigg(
				r_i \E{q(\mu)}\Big[\ln \mu_j\Big] + 
				(1 -r_i) \E{q(\mu)}\Big[\ln (1-\mu_j)\Big]
			\Bigg)\\	
		=&
			\sum_{i=1}^N \sum_{j=1}^K
			\phi_i(j) \Bigg(
				r_i \Big( \psi(\alpha'_{1,j}) - \psi(\alpha'_{1,j} + \alpha'_{0,j}) \Big) + 
				(1 -r_i) \Big( \psi(\alpha'_{0,j}) - \psi(\alpha'_{1,j} + \alpha'_{0,j}) \Big)
			\Bigg)\\
		=&
			\sum_{i=1}^N \sum_{j=1}^K
			\phi_i(j) \Bigg(
				r_i \psi(\alpha'_{1,j}) + 
				(1 -r_i) \psi(\alpha'_{0,j}) - 
				\psi(\alpha'_{1,j} + \alpha'_{0,j})
			\Bigg)\\	
\end{aligned}
\end{equation}
Second expectation:
\begin{equation}
\begin{aligned}
	\E{q(a)q(\mu)q(\pi)} \Bigg[\sum_{i=1}^N \sum_{j=1}^K \sum_{k=1}^K \ictr[z_i = k] \ln p(a_i = j | \pi_k)\Bigg]
		=&
			\sum_{i=1}^N \sum_{j=1}^K \sum_{k=1}^K \ictr[z_i = k] 
			\E{q(a)q(\mu)q(\pi)} \Bigg[ \ln (\pi_{k,j}^{\ictr[a_i = j]}) \Bigg]\\
		=&
			\sum_{i=1}^N \sum_{j=1}^K \sum_{k=1}^K \ictr[z_i = k]
			\E{q(a)q(\mu)q(\pi)} \Bigg[ \ictr[a_i = j] \ln \pi_{k,j} \Bigg]\\
		=&
			\sum_{i=1}^N \sum_{j=1}^K \sum_{k=1}^K \ictr[z_i = k]
			\E{q(a)} \Big[ \ictr[a_i = j] \Big]
			\E{q(\pi)} \Big[ \ln \pi_{k,j} \Big]\\
		=&
			\sum_{i=1}^N \sum_{j=1}^K \sum_{k=1}^K \ictr[z_i = k]
			\phi_i(j)
			\Bigg( \psi(\beta'_{k,j}) - \psi\Big(\sum_{l=1}^K \beta'_{k,l} \Big) \Bigg)\\
\end{aligned}
\end{equation}
Third expectation:
\begin{equation}
\begin{aligned}
	\E{q(a)q(\mu)q(\pi)} \Bigg[\sum_{j=1}^K \ln p(\mu_j)\Bigg]
		=&
			\sum_{j=1}^K \E{q(\mu)} \Bigg[\ln \Big( (\mu_j)^{(\alpha_1-1)}(1 - \mu_j)^{(\alpha_0-1)} \Big) \Bigg]\\
		=&
			\sum_{j=1}^K \E{q(\mu)} \Bigg[(\alpha_1-1) \ln (\mu_j) + (\alpha_0-1) \ln (1 - \mu_j) \Bigg]\\
		=&
			\sum_{j=1}^K \Bigg[
				(\alpha_1-1)\E{q(\mu)} \Big[ \ln (\mu_j) \Big]+ 
				(\alpha_0-1)\E{q(\mu)} \Big[ \ln (1 - \mu_j) \Big]
			\Bigg]\\
		=&
			\sum_{j=1}^K \Bigg[
				(\alpha_1-1) \Big( \psi(\alpha'_{1,j}) - \psi(\alpha'_{1,j} + \alpha'_{0,j}) \Big)+ 
				(\alpha_0-1) \Big( \psi(\alpha'_{0,j}) - \psi(\alpha'_{1,j} + \alpha'_{0,j}) \Big)
			\Bigg]\\
\end{aligned}
\end{equation}
Fourth expectation:
\begin{equation}
\begin{aligned}
	\E{q(a)q(\mu)q(\pi)} \Bigg[\sum_{k=1}^K \ln p(\pi_k)\Bigg]
		=&
			\sum_{k=1}^K \E{q(a)q(\mu)q(\pi)} \Bigg[\ln \Bigg( 
				\frac{1}{B(\beta)} \prod_{l=1}^K \pi_{k,l}^{(\beta_l - 1)}				
			\Bigg) \Bigg]\\
		=&
			\sum_{k=1}^K \E{q(\pi)} \Bigg[
				\ln \Bigg(\prod_{l=1}^K \pi_{k,l}^{(\beta_l - 1)} \Bigg) - \ln B(\beta) 				
			\Bigg]\\
		=&
			\sum_{k=1}^K \E{q(\pi)} \Bigg[
				\sum_{l=k}^K (\beta_l - 1) \ln \pi_{k,l} - \ln B(\beta) 				
			\Bigg]\\
		=&
			\sum_{k=1}^K 
				\sum_{l=k}^K (\beta_l - 1) \E{q(\pi)} \Big[\ln \pi_{k,l} \Big] -
				\ln B(\beta) 				
			\\
		=&
			\sum_{k=1}^K \sum_{l=k}^K 
				 (\beta_l - 1) \Bigg( \psi(\beta'_{k,l}) - \psi\Big(\sum_{m=1}^K \beta'_{k,m} \Big) \Bigg) +
				 \const\\
\end{aligned}
\end{equation}
Fifth expectation (entropy):
\begin{equation}
\begin{aligned}
	-\E{q(a)} \Big[\sum_{i=1}^N  \ln q(a_i) \Big]
		=&
			-\sum_{i=1}^N \E{q(a)} \Big[ \ln q(a_i) \Big]\\
		=&
			-\sum_{i=1}^N \sum_{j=1}^{K} \phi_i(j) \ln \phi_i(j)\\
\end{aligned}
\end{equation}
Sixth expectation (entropy):
\begin{equation}
\begin{aligned}
	-\E{q(\mu)} \Bigg[ &\sum_{j=1}^K \ln q(\mu_j) \Bigg]\\
		=&
			\sum_{j=1}^K -\E{q(\mu)} \Bigg[\ln q(\mu_j) \Bigg]\\
		=& 
			\sum_{j=1}^K 
				\ln B(\alpha'_{1,j}, \alpha'_{0,j})-
				(\alpha'_{1,j} - 1)\psi(\alpha'_{1,j}) - 
				(\alpha'_{0,j} - 1)\psi(\alpha'_{0,j}) +
				(\alpha'_{1,j} + \alpha'_{0,j} - 2) \psi(\alpha'_{1,j} + \alpha'_{0,j})\\
		=& 
			\sum_{j=1}^K 
				\ln \Bigg(
					\frac{\Gamma(\alpha'_{1,j})\Gamma(\alpha'_{0,j})}
					{\Gamma(\alpha'_{1,j} + \alpha'_{0,j})}
				\Bigg) - 
				(\alpha'_{1,j} - 1)\psi(\alpha'_{1,j}) - 
				(\alpha'_{0,j} - 1)\psi(\alpha'_{0,j}) +
				(\alpha'_{1,j} + \alpha'_{0,j} - 2) \psi(\alpha'_{1,j} + \alpha'_{0,j})\\
\end{aligned}
\end{equation}
Seventh expectation (entropy):
\begin{equation}
\begin{aligned}
	-\E{q(\pi)} \Bigg[ \sum_{k=1}^K \ln q(\pi_k) \Bigg]
		=&
			\sum_{k=1}^K -\E{q(\pi)} \Bigg[\ln q(\pi_k) \Bigg]\\
		=&
			\sum_{k=1}^K
				\ln B(\beta'_{k}) - 
				(K - \beta'_{k,0}) \psi(\beta'_{k,0}) - 
				\sum_{l=1}^K (\beta'_{k,l} - 1) \psi(\beta'_{k,l})\\
		=&
			\sum_{k=1}^K
				\ln \Bigg(
					\frac{\prod_{l=1}^K \Gamma(\beta'_{k,l})}
					{\Gamma\Big(\sum_{l=1}^K \beta'_{k,l}\Big)}
				\Bigg) - 
				\Big(K - \sum_{l=1}^K \beta'_{k,l}\Big) \psi\Big(\sum_{l=1}^K \beta'_{k,l}\Big) - 
				\sum_{l=1}^K (\beta'_{k,l} - 1) \psi(\beta'_{k,l})\\
\end{aligned}
\end{equation}
We now assemble these seven results to present the variational objective in its full glory:
\begin{equation}
\begin{aligned}
	\mathcal{L}(\phi, &\alpha', \beta')=\\
		&
			\sum_{i=1}^N \sum_{j=1}^K
			\phi_i(j) \Bigg(
				r_i \psi(\alpha'_{1,j}) + 
				(1 -r_i) \psi(\alpha'_{0,j}) - 
				\psi(\alpha'_{1,j} + \alpha'_{0,j})
			\Bigg) + \\
		&
			\sum_{i=1}^N \sum_{j=1}^K \sum_{k=1}^K \ictr[z_i = k]
			\phi_i(j)
			\Bigg( \psi(\beta'_{k,j}) - \psi\Big(\sum_{l=1}^K \beta'_{k,l} \Big) \Bigg) + \\
		&
			\sum_{j=1}^K \Bigg[
				(\alpha_1-1) \Big( \psi(\alpha'_{1,j}) - \psi(\alpha'_{1,j} + \alpha'_{0,j}) \Big)+ 
				(\alpha_0-1) \Big( \psi(\alpha'_{0,j}) - \psi(\alpha'_{1,j} + \alpha'_{0,j}) \Big)
			\Bigg] + \\
		&	
			\sum_{k=1}^K \sum_{l=k}^K 
				 (\beta_l - 1) \Bigg( \psi(\beta'_{k,l}) - \psi\Big(\sum_{m=1}^K \beta'_{k,m} \Big) \Bigg) - \\
		&
			\sum_{i=1}^N \sum_{j=1}^{K} \phi_i(j) \ln \phi_i(j) + \\
		&
			\sum_{j=1}^K 
				\ln \Bigg(
					\frac{\Gamma(\alpha'_{1,j})\Gamma(\alpha'_{0,j})}
					{\Gamma(\alpha'_{1,j} + \alpha'_{0,j})}
				\Bigg) - 
				(\alpha'_{1,j} - 1)\psi(\alpha'_{1,j}) - 
				(\alpha'_{0,j} - 1)\psi(\alpha'_{0,j}) +
				(\alpha'_{1,j} + \alpha'_{0,j} - 2) \psi(\alpha'_{1,j} + \alpha'_{0,j}) + \\
		&
			\sum_{k=1}^K
				\ln \Bigg(
					\frac{\prod_{l=1}^K \Gamma(\beta'_{k,l})}
					{\Gamma\Big(\sum_{l=1}^K \beta'_{k,l}\Big)}
				\Bigg) - 
				(K - \beta'_{k,0}) \psi(\beta'_{k,0}) - 
				\sum_{l=1}^K (\beta'_{k,l} - 1) \psi(\beta'_{k,l}) + \\
		&	\const
\end{aligned}
\end{equation}
With the variational parameter updates and the variational objective derived, we can now put forth the Variational Inference algorithm that approximates the posterior, $p(a, \mu, \pi| z, r)$, needed by TS-Lat from \cref{sec:ts_lat}.

\begin{algorithm}[H]
   \caption{Variational Inference for MAB with Latent Noncompliance}
   \label{alg:vi}
\begin{algorithmic}
	\STATE Initialize: $\phi_i(j) \sim \Dir{\beta} \ \forall \ i \in \{1,\hdots,N\}, j \in \{1,\hdots,K\}$
	\STATE converged = false
	\STATE $\mathcal{L}_{t-1}(\phi, \alpha', \beta') = -\infty$
	\WHILE{\text{not converged}}
	    \STATE Update $q(\mu)$ variational parameters $\forall \ j \in \{1,\hdots,K\}$:
	    \begin{equation*}
	    \begin{aligned}
	    	\alpha'_{1,j} := 
	    	\alpha_1 + \sum\limits_{i=1}^N r_i\phi_i(j) \ \ \text{ and } \ \ 
			\alpha'_{0,j} := 
			\alpha_0 + \sum\limits_{i=1}^N (1 -r_i)\phi_i(j)
	    \end{aligned}
	    \end{equation*}
	    \STATE Update $q(\pi)$ variational parameters $k \in \{1,\hdots,K\}, l\in \{1,\hdots,K\}$:
	    \begin{equation*}
	    \begin{aligned}
	    	\beta'_{k,l}
				:=& \beta + \sum_{i=1}^N \ictr [z_i = k] \cdot \phi_i(l)\\
	    \end{aligned}
	    \end{equation*}
	    \STATE Update $q(a)$ variational parameters $\forall \ i \in \{1,\hdots,N\}, j \in \{1,\hdots,K\}$:
	    {\small \begin{equation*}
	    \begin{aligned}
	    	\phi_i(j) =& 
	    		\frac{\exp\Bigg(
			r_i \Big(\psi(\alpha'_{1,j}) - \psi(\alpha'_{1,j} + \alpha'_{0,j})\Big)+
			(1-r_i) \Big(\psi(\alpha'_{0,j}) - \psi(\alpha'_{1,j} + \alpha'_{0,j})\Big) +
			\psi(\beta'_{z_i,j}) - \psi\Big(\sum_{l=1}^K \beta'_{z_i,l}\Big)
		\Bigg)}{\sum\limits_{k=1}^K \exp\Bigg(
			r_i \Big(\psi(\alpha'_{k,1}) - \psi(\alpha'_{k,1} + \alpha'_{k,0})\Big)+
			(1-r_i) \Big(\psi(\alpha'_{k,0}) - \psi(\alpha'_{k,1} + \alpha'_{k,0})\Big) +
			\psi(\beta'_{z_i,k}) - \psi\Big(\sum_{l=1}^K \beta'_{z_i,l}\Big)
		\Bigg)}\\
	    \end{aligned}
	    \end{equation*}}
	    \STATE Compute: $\mathcal{L}_{t}(\phi, \alpha', \beta')$
	    \IF{$\mathcal{L}_{t}(\phi, \alpha', \beta') - \mathcal{L}_{t-1}(\phi, \alpha', \beta') < \epsilon$}
	    		\STATE converged = true
	    \ENDIF
	    \STATE $\mathcal{L}_{t-1}(\phi, \alpha', \beta') = \mathcal{L}_{t}(\phi, \alpha', \beta')$
	\ENDWHILE
	\RETURN $(\alpha', \beta')$
\end{algorithmic}
\end{algorithm}

\end{document}